\documentclass{article}

\usepackage[english]{babel}

\usepackage[letterpaper,top=2cm,bottom=2cm,left=3cm,right=3cm,marginparwidth=1.75cm]{geometry}

\usepackage{amsmath}
\usepackage{graphicx}
\usepackage{authblk}
\usepackage[labelformat=empty, position=top]{subcaption}

\usepackage[citestyle=numeric, bibstyle=authoryear, giveninits=true, sorting=none]{biblatex}
\usepackage{xpatch}
\DeclareNameAlias{sortname}{last-first}

\renewbibmacro*{in:}{%
    \iffieldequalstr{entrytype}{inproceedings}{%
        \printtext{\bibstring{in}\addspace}%
    }{}%
}
\csletcs{abx@macro@publisher+location+date@orig}{abx@macro@publisher+location+date}
\renewbibmacro*{publisher+location+date}{%
    \printtext[parens]{\usebibmacro{publisher+location+date@orig}}
}
\DeclareFieldFormat[book]{title}{#1\printunit{\addspace}}
\DeclareFieldFormat[inproceedings]{title}{#1\isdot}
\DeclareFieldFormat{booktitle}{#1\addcomma}
\xpatchbibmacro{byeditor+others}{%
    \usebibmacro{byeditor+othersstrg}%
    \setunit{\addspace}%
    \printnames[byeditor]{editor}%
    \clearname{editor}%
}{%
    \printnames[byeditor]{editor}%
    \clearname{editor}
    \addcomma\addspace
    \bibstring{editor}
    \setunit{\addspace}%
}{}{}
\DeclareFieldFormat[article]{title}{#1}
\DeclareFieldFormat[article]{journaltitle}{#1\isdot}
\DeclareFieldFormat[article]{volume}{#1}
\DeclareFieldFormat[article]{pages}{#1}

\makeatletter
\input{numeric.bbx}
\makeatother

\AtEveryBibitem{%
  \clearfield{number}}

\DeclareNameAlias{author}{last-first}

\title{The Relational Bottleneck as an Inductive Bias\\for Efficient Abstraction}
\author[1,*]{Taylor W. Webb}
\author[2]{Steven M. Frankland}
\author[3]{Awni Altabaa}
\author[4]{Simon Segert}
\author[4]{Kamesh Krishnamurthy}
\author[4]{Declan Campbell}
\author[5]{Jacob Russin}
\author[4]{Tyler Giallanza}
\author[4]{Zack Dulberg}
\author[6]{Randall O’Reilly}
\author[3]{John Lafferty}
\author[4]{Jonathan D. Cohen}
\affil[1]{University of California, Los Angeles}
\affil[2]{Dartmouth College}
\affil[3]{Yale University}
\affil[4]{Princeton University}
\affil[5]{Brown University}
\affil[6]{University of California, Davis}
\affil[*]{Correspondence to: taylor.w.webb@gmail.com}

\date{}

\begin{document}
\maketitle

\begin{abstract}
A central challenge for cognitive science is to explain how abstract concepts are acquired from limited experience. This has often been framed in terms of a dichotomy between connectionist and symbolic cognitive models. Here, we highlight a recently emerging line of work that suggests a novel reconciliation of these approaches, by exploiting an inductive bias that we term the \textit{relational bottleneck}. In that approach, neural networks are constrained via their architecture to focus on relations between perceptual inputs, rather than the attributes of individual inputs. We review a family of models that employ this approach to induce abstractions in a data-efficient manner, emphasizing their potential as candidate models for the acquisition of abstract concepts in the human mind and brain.
\end{abstract}

\section*{Highlights}

\begin{itemize}
\item Human learners acquire abstract concepts from limited experience. The effort to explain this capacity has fueled debate between symbolic and connectionist approaches, and motivated proposals for neuro-symbolic systems.
\item The \textit{relational bottleneck} principle suggests a novel way to bridge the gap. By restricting information processing to focus only on relations, the approach encourages abstract symbol-like mechanisms to emerge in neural networks.
\item We present an information theoretic formulation, and review neural network architectures that implement the principle, enabling rapid learning and systematic generalization of relational patterns.
\item The approach can explain phenomena ranging from the development of numerical abstractions to capacity limits in cognition; is consistent with findings from cognitive neuroscience; and offers a principle for designing more powerful artificial learning systems.
\end{itemize}


\section*{Modeling the efficient induction of abstractions}

Human cognition displays a remarkable ability to transcend the specifics of limited experience to entertain highly general, abstract ideas. Understanding how the mind and brain accomplish this has been a central challenge of cognitive science, and a major preoccupation of philosophy before that~\cite{descartes1628rules, locke1894essay, leibniz1996essays, chomsky1980review}. Of particular importance is the central role played by \textit{relations}, which enable human reasoners to abstract away from individual entities and identify higher-order patterns across distinct domains~\cite{gentner1983structure,holyoak2012analogy}. For instance, when presented with the images in Figure~\ref{rel_bottleneck_illustrated}, one can easily determine that a common relational pattern (ABA) is displayed on both the left and right, despite the involvement of completely different objects. This capacity is a major component underlying the human capacity for fluid reasoning~\cite{cattell1971abilities,snow1984topography}, and has been proposed as a key factor distinguishing human intelligence from that of other species~\cite{penn2008darwin}.

A long tradition in cognitive science and AI~\cite{newell1972human,fodor1975language,anderson1996act} holds that this capacity for abstraction depends on processes akin to symbolic programs. A major appeal of this approach is that symbols are, by design, abstracted away from the content to which they refer, thus naturally accounting for the flexibility and systematicity of human concepts~\cite{fodor1988connectionism}. More recently, \textbf{program induction} (see Glossary) has provided an account of how symbolic concepts might be learned directly from data~\cite{lake2015human,lake2017building,rule2020child,ellis2021dreamcoder,dehaene2022symbols,yang2022one,quilty2022best}, formalizing learning as a search for the program that maximizes the likelihood of observed data. However, while this approach is capable in principle of representing any possible set of concepts~\cite{piantadosi2021computational}, the discovery of these concepts using traditional search methods often proves intractable, making it difficult in practice to identify programs with the richness and complexity of human natural concepts (though see Box 1 for discussion).

An alternative approach, \textbf{connectionism}, has for decades explored how cognitive abstractions might emerge through experience in general-purpose neural architectures~\cite{mcclelland1989explorations,elman1990finding,mcclelland2003parallel,mcclelland2010letting}. This endeavor has taken on new relevance with the advent of large language models, demonstrating that it is possible, in some cases, for a human-like capacity for abstraction to emerge given sufficient scaling of architecture and training data~\cite{brown2020language,wei2022emergent,piantadosi2023modern,bubeck2023sparks}. For instance, it has recently been shown that large language models can solve various analogy problems at a level equal to that of college students~\cite{webb2023emergent}. However, the ability of these models to perform abstract tasks depends on exposure to a much larger training corpus than individual humans receive in an entire lifetime~\cite{griffiths2020understanding, frank2023bridging}, thus failing to account for the data efficiency of human concept learning.

In this review, we highlight an emerging approach that suggests a novel reconciliation of these two traditions. The central feature of this approach is an \textbf{inductive bias} that we refer to as the \textit{relational bottleneck}: a constraint that biases neural network models to focus on relations between objects rather than the attributes of individual objects. This approach enables the data efficiency associated with symbolic cognitive models, while retaining the scalable training procedures associated with neural network models (see Box 1 for further discussion of neuro-symbolic approaches). In the following sections, we first provide a general characterization of this approach, drawing on information theory, and discuss recently proposed neural network architectures that implement the approach. We then discuss the potential of the approach for modeling human cognition, relating it to existing theories and considering mechanisms through which it might be implemented in the brain. 


\noindent\rule{\textwidth}{1pt}
\subsubsection*{Box 1: Neuro-symbolic modeling approaches}

Many approaches have been proposed for hybrid systems that combine aspects of both neural and symbolic computing. Early work in this area focused on incorporating a capacity for variable-binding – a key property of symbolic systems – into connectionist systems. Notable examples of this approach include binding-by-synchrony~\cite{hummel1997distributed}, tensor product variable-binding~\cite{smolensky1990tensor}, and BoltzCONS~\cite{touretzky1990boltzcons}. A number of vector symbolic architectures have since been proposed that build on the tensor product operation, but enable more elaborate symbolic structures to be embedded in a vector space of fixed dimensionality~\cite{plate1995holographic,kanerva2009hyperdimensional,eliasmith2012large,schlegel2022comparison}. These approaches have all generally relied on the use of pre-specified symbolic primitives.

More recently, hybrid systems have been developed that combine deep learning with symbolic programs. In this approach, deep learning components are typically employed to translate raw perceptual inputs, such as images or natural language, into symbolic representations, which can then be processed by traditional symbolic algorithms~\cite{johnson2017inferring,yi2018neural,nye2020learning}. This approach is complemented by recent neuro-symbolic approaches to probabilistic program induction, in which symbolic primitives are pre-specified (following earlier symbolic-connectionist modeling efforts), and then deep learning is used to assemble these primitives into programs~\cite{ellis2021dreamcoder}.   

An alternative approach (which might also be viewed as neuro-symbolic in some sense) involves the integration of key features of symbolic computing within the framework of end-to-end trainable neural systems. Examples of this approach include neural production systems~\cite{goyal2021neural}, graph neural networks~\cite{battaglia2018relational}, discrete-valued neural networks~\cite{liu2021discrete} (see~\cite{feldman2012symbolic} for further discussion of normative considerations regarding discrete-valued representations), sparse causal graphs~\cite{ke2019learning}, and efforts to incorporate tensor product representations into end-to-end systems~\cite{palangi2018question,jiang2021enriching}. The relational bottleneck falls into this broad category, as it incorporates key elements of symbolic computing – variable-binding and relational representations – into fully differentiable neural systems that can be trained end-to-end without the need for pre-specified symbolic primitives. Relative to these other approaches, the primary innovation of the relational bottleneck framework is the emphasis on architectural components that promote the development of genuinely relational representations.

\noindent\rule{\textwidth}{1pt}


\section*{The relational bottleneck}

\begin{figure}[h!]
\centering
\includegraphics[width=0.85\linewidth]{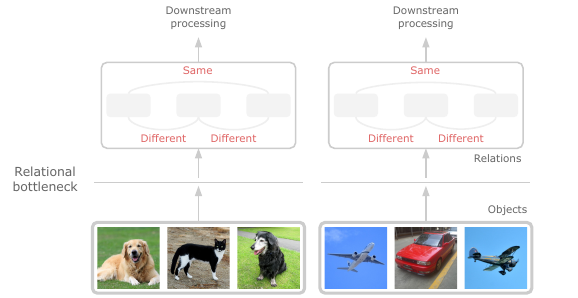} 
\caption{\textbf{The relational bottleneck.} An inductive bias that prioritizes the representation of relations (e.g., `same' vs. `different'), and discourages the representation of the features of individual objects (e.g., the shape or color of the objects in the images above). The result is that downstream processing is driven primarily, or even exclusively by patterns of relations, and can therefore systematically generalize those patterns across distinct instances (e.g., the common ABA pattern displayed on both left and right), even for completely novel objects. The approach is illustrated here with same/different relations, but other relations can also be accommodated. Note that this example is intended only to illustrate the overall goal of the relational bottleneck framework. Figure~\ref{rel_bottleneck_architectures} depicts neural architectures that implement the approach.}  
\label{rel_bottleneck_illustrated}
\end{figure}

We define the relational bottleneck as any mechanism that restricts the flow of information from perceptual to downstream reasoning systems to consist only of relations (see Box 2 for a formal definition). For example, consider the relational patterns depicted in Figure~\ref{rel_bottleneck_illustrated}. In this example, the images on the left and right are governed by the same abstract pattern (ABA), but these patterns contain different sets of objects (dogs and cats vs. planes and cars). Given inputs representing individual objects (e.g., representations of the dog and cat images on the left), a relational bottleneck would constrain the representations passed to downstream reasoning processes to capture only the relations between these objects (e.g., whether the objects have the same shape), and discard information about the individual objects (e.g., information about dogs and cats). The result is that the images on the left will have the same representation as those on the right, despite the different objects depicted in these images. This encourages downstream processes to identify relational patterns in a manner that is abstracted away from specific instances of those patterns, and can therefore be systematically generalized to novel inputs (see Concluding Remarks for discussion of non-relational factors in human cognition). In the following section, we highlight three recently proposed neural architectures that instantiate this approach in different guises, illustrating how they utilize a relational bottleneck to induce abstract concepts in a data-efficient manner. It is also worth emphasizing that, although we focus here on artificial neural networks, the approach may also be applicable to other modeling approaches (see Outstanding Questions).



\noindent\rule{\textwidth}{1pt}
\subsubsection*{Box 2: The relational bottleneck principle}

Information bottleneck theory~\cite{tishby2000information} provides a normative framework for formalizing the notion of a relational bottleneck. Consider an information processing system that receives an input signal $X$ and aims to predict a target signal $Y$. $X$ is processed to generate a compressed representation $Z = f(X)$ (the `bottleneck'), which is then used to predict $Y$. At the heart of information bottleneck theory is the idea of `minimal-sufficiency'. $Z$ is \textit{sufficient} for predicting $Y$ if it contains all the information $X$ encodes about $Y$. That is, $I(Z; Y) = I(X; Y)$, where $I(\cdot\,; \,\cdot)$ is the mutual information. If $Z$ is sufficient, then we write $X \to Z \to Y$, meaning that $Y$ is conditionally independent of $X$ given the compressed representation $Z$. $Z$ is \textit{minimal-sufficient} if it is sufficient for $Y$ and does not contain any extraneous information about $X$ which is not relevant to predicting $Y$. That is, $I(X; Z) \leq I(X; \tilde{Z})$ for any other sufficient compressed representation $\tilde{Z}$.

Achieving maximum compression while retaining as much relevant information as possible is a trade-off. It is captured by the information bottleneck objective,
\begin{equation}
\label{eq:info_bottleneck_objective}
    \mathrm{minimize} \ \mathcal{L}(Z) = I(X; Z) - \beta I(Z; Y).
\end{equation}
 
This objective reflects the tension between compression -- which favors discarding information as captured by the first term -- and the preservation of relevant information in $Z$, captured by the second term. The parameter $\beta$ controls this trade-off. 

While this objective is well-defined when the joint distribution $(X, Y)$ is known, obtaining a minimal-sufficient compressed representation from data is, in general, very challenging for the high-dimensional signals that are often of interest. However, it may be possible to implicitly enforce a desirable information bottleneck for a large class of tasks through architectural inductive biases.

In particular, we hypothesize that human cognition has been fundamentally optimized for tasks that are relational in nature. We define a `relational task' as any task for which there exists a minimal-sufficient representation $R$ that is \textit{purely relational}. Suppose the input signal represents a set of objects,
\begin{equation}
    X = \left( x_{1},\, \ldots,\, x_{N}\right).
\end{equation}

A relational signal is a signal of the form,
\begin{equation}
    R = \left\{r(x_i, x_j) \right\}_{i\neq j} = \left\{ r(x_1, x_2),\, r(x_1, x_3),\, \ldots,\, r(x_{N-1}, x_{N}) \right\},
\end{equation}
where $r(x_i, x_j)$ is a relation function. A ``relation function'' is a function that takes a pair of objects as input and returns a relation between them (see `Modeling more complex relations' for discussion of higher-order relations). We say that a task is relational if there exists some relational signal $R$ which is sufficient for predicting the target $Y$ (i.e., $X \to R \to Y$). A relational bottleneck is any mechanism that restricts or biases the learned compressed representation of the input to be a relational representation, separated from object-level features. This gives the model a smaller space of possible compressed representations over which it must search. Moreover, this restricted space is guaranteed to contain a sufficient representation for the task and excludes many representations that encode extraneous information about $X$, promoting efficient learning of relational abstractions.

In practice, a relational bottleneck can be implemented in a learning model through the use of architectural inductive biases. One particularly useful operation is inner products of the form $\langle \phi(x_i), \phi(x_j) \rangle$, which naturally capture a notion of relations in terms of similarity between learned attributes. Inner products can also capture asymmetric relations through the use of separate encoders for $x_i$ and $x_j$ (i.e., $\langle \phi(x_i), \psi(x_j) \rangle$), and indeed can be shown to be universal approximators~\cite{altabaa2024approximation}. The class of functions that can be represented in this way is thus fully general, but the approach provides a useful inductive bias for disentangling feature extraction (implemented by the encoders $\phi$ and $\psi$) from comparison (implemented by the inner product operator $\langle \rangle$).

\noindent\rule{\textwidth}{1pt}

\section*{The relational bottleneck in neural architectures}

\begin{figure}[h!]
\centering
\begin{subfigure}[t]{.1\linewidth}\vskip 0pt
    \subcaption{}
    \label{ESBN_arch}
\end{subfigure}
\begin{subfigure}[t]{.1\linewidth}\vskip 0pt
    \subcaption{}
    \label{corelnet_arch}
\end{subfigure}
\begin{subfigure}[t]{.1\linewidth}\vskip 0pt
    \subcaption{}
    \label{abstractor_arch}
\end{subfigure}
\includegraphics[width=0.9\linewidth]{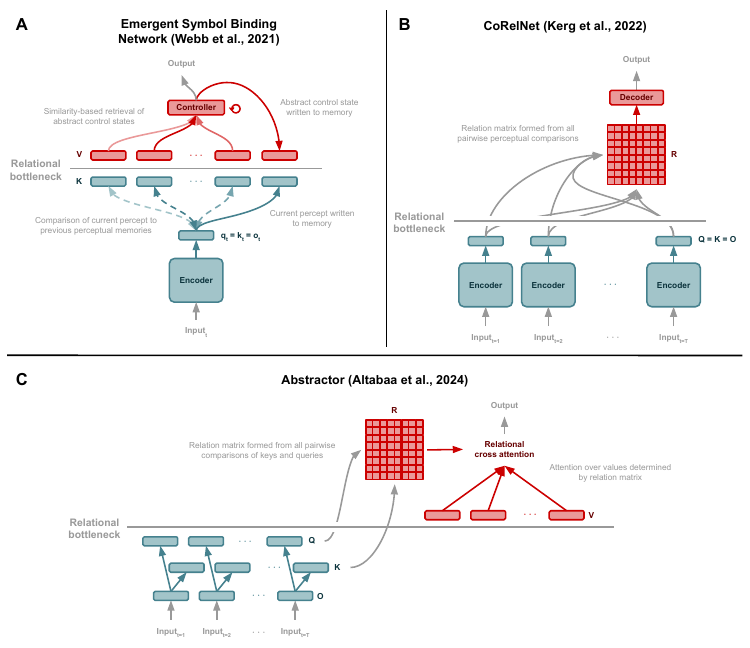} 
\caption{\textbf{Implementing the relational bottleneck.} Three neural architectures that implement the relational bottleneck. \textbf{(a)} Emergent Symbol Binding Network (ESBN)~\cite{webb2020emergent}. \textbf{(b)} Compositional Relation Network (CoRelNet)~\cite{kerg2022neural}. \textbf{(c)} Abstractor~\cite{altabaa2023abstractors}. In all cases, high-dimensional inputs (e.g., images) are processed by a neural encoder (e.g., a convolutional network), yielding a set of object embeddings $\mathbf{O}$. These are projected to a set of keys $\mathbf{K}$ and queries $\mathbf{Q}$, which are then compared yielding a relation matrix $\mathbf{R}$, in which each entry is an inner product between a query and key. Abstract values $\mathbf{V}$ are isolated from perceptual inputs (the core feature of the relational bottleneck), and depend only on the relations between them.}  
\label{rel_bottleneck_architectures}
\end{figure}

Figure~\ref{rel_bottleneck_architectures} (Key Figure) depicts three architectures that implement the relational bottleneck through architectural inductive biases. Here, we discuss how the distinct mechanisms in these models implement the same underlying principle. In particular, a common element is the use of \textbf{inner products} to represent relations, which ensures that the resulting representations are genuinely relational. In each case, we also contrast these architectures with related approaches that do \textit{not} incorporate a relational bottleneck, emphasizing how this key architectural feature enables the data-efficient induction of abstractions.

\subsection*{Emergent symbol binding}

We first consider the Emergent Symbol Binding Network (ESBN) (Figure~\ref{ESBN_arch})~\cite{webb2020emergent}, a deep neural network architecture inspired by the notion of role-filler variable binding in cognitive models of relational reasoning~\cite{smolensky1990tensor,hummel1997distributed,marcus2003algebraic}. In those models, relational reasoning is supported by the existence of separate `roles', representing information about abstract variables, and `fillers', representing information about concrete entities bound to those variables. This coding scheme enables roles and fillers to be flexibly combined in new ways, capturing a key property of symbol-processing systems: the ability of symbolic variables to be associated with any potential values. Previous work focused on how binding might be performed in neural circuits (e.g., units coding for a particular role may be temporarily `bound' to units coding for a particular filler by firing synchronously~\cite{hummel1997distributed}). However, role and filler representations were typically pre-specified by the modeler, leaving open the question of how such symbolic representations might be learned.

The ESBN adopts this key idea of separate roles and fillers, but integrates them into a system that can be trained end-to-end (via \textbf{backpropagation}), averting the need to pre-specify these representations. The ESBN contains three major components: 1) a feedforward encoding pathway (`Encoder' in Figure~\ref{ESBN_arch}), which generates object \textbf{embeddings} (i.e., fillers) from perceptual inputs (e.g., images); 2) a recurrent controller (`Controller' in Figure~\ref{ESBN_arch}), which operates over learned representations of abstract variables (i.e., roles; though the system is not explicitly trained to represent any particular variables, but instead learns these representations through backpropagation); and 3) an \textbf{external memory} system responsible for binding representations of roles and fillers (i.e., abstract variables and perceptual embeddings). The ESBN processes inputs sequentially. For each observation, a pair of representations is appended to memory, one from the perceptual pathway (referred to as a \textit{key}), and one from the control pathway (referred to as a \textit{value}; note that the use of the terms `key' and `value' here is reversed relative to the original paper~\cite{webb2020emergent} in order to be consistent with their usage in describing the CoRelNet and Abstractor architectures). To read from this memory, the embedding for the current observation (referred to as a \textit{query}) is compared to all keys in memory via an inner product, yielding a set of scores (one for each key) that indicate the similarity of the current object embedding (i.e., filler) to each entry in memory. These scores are then used to compute a weighted average of the value embeddings (i.e., roles) in the abstract pathway, which is then retrieved and passed to the controller.

\begin{figure}[h!]
\centering
\begin{subfigure}[t]{.1\linewidth}\vskip 0pt
    \subcaption{}
    \label{ESBN_results}
\end{subfigure}
\begin{subfigure}[t]{.1\linewidth}\vskip 0pt
    \subcaption{}
    \label{abstractor_results}
\end{subfigure}
\begin{subfigure}[t]{.1\linewidth}\vskip 0pt
    \subcaption{}
    \label{counting_results}
\end{subfigure}
\includegraphics[width=\linewidth]{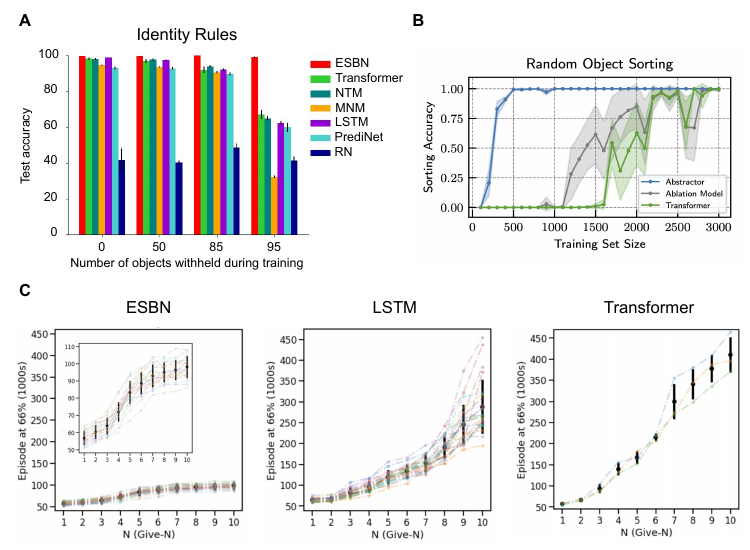} 
\caption{\textbf{The relational bottleneck encourages data-efficient and generalizable relation learning.} \textbf{(a)} Results for the ESBN and baseline architectures (Transformer, Neural Turing Machine (NTM), Metalearned Neural Memory (MNM), Long Short-Term Memory (LSTM), PrediNet, and the Relation Network (RN)) on the identity rules task, reproduced from~\cite{webb2020emergent}. X axis represents the number of potential objects (out of 100 possible objects) withheld during training. When all objects are observed during training (0 withheld), most baselines perform well on the task. When most objects are withheld (95 withheld; test set includes only objects withheld during training), only the ESBN generalizes well to new objects. \textbf{(b)} Results for an object sorting task involving an asymmetric relation (greater-than/less-than), reproduced from~\cite{altabaa2023abstractors}. The abstractor learns this task significantly faster than both the transformer and an ablation model in which relational cross-attention is replaced by standard cross-attention. \textbf{(c)} Results from the give-N task, reproduced from~\cite{dulberg2021modelling}. X axis represents the target number N (desired number of objects). Y axis represents the episode at which the model reaches a particular criterion for the ability to count to each value of N. ESBN learns the task significantly faster than the LSTM or Transformer baselines. ESBN also displays inductive transition (rapid learning for $N>5$) similar to that observed in human development.}  
\label{rel_bottleneck_results}
\end{figure}

Importantly, in this retrieval operation, the control pathway cannot access the \textit{content} of the representations in the perceptual pathway (i.e., the fillers). Instead, the interaction is mediated only by the \textit{comparison} of perceptual representations with each other. In other words, to the extent that a particular perceptual memory embedding (key) is similar to the current perceptual embedding (filler, used as a query), the corresponding control memory (value) will be retrieved and passed to the controller to be used as the role; critically, the perceptual embedding itself is not passed to the controller. The ESBN thus implements the relational bottleneck as an architectural prior, separating the learning and use of abstract representations (roles) by the controller from the embeddings of perceptual information (fillers). It is precisely this separation that guarantees the representations in the control pathway are abstract. This feature enables the ESBN to rapidly learn relational patterns (such as the identity rules displayed in Figure~\ref{rel_bottleneck_illustrated}), and generalize them to \textbf{out-of-distribution} inputs (e.g., previously unseen shapes; Figure~\ref{ESBN_results})~\cite{webb2020emergent}. Indeed, the ESBN can learn relational patterns from only a handful of examples (as few as 4 examples), mirroring the data efficiency of relational learning in young children (who can typically learn to perform relational tasks with fewer than $\sim20$ examples~\cite{kotovsky1996comparison}, as opposed to the thousands of examples that are necessary for standard neural network architectures). Critically, the ESBN can be shown to use precisely the same representation for a given role, irrespective of filler, thus exhibiting a critical feature of abstract, symbolic processing~\cite{fodor1988connectionism}. In this sense, the representations in the model's control pathway can be viewed as a form of learned `symbols'.

It is instructive to compare this model to similar approaches without a relational bottleneck. The ESBN is part of a broader family of architectures that use content-addressable \textbf{external memory} -- a separate store of information with which a neural network can interact via learnable read and write operations~\cite{graves2014neural,graves2016hybrid}. Notably, these read and write operations rely on a similarity computation (based on inner products). These have often been cast as simplified implementations of the brain's episodic memory system~\cite{tulving2002episodic,mcclelland1995there}. Standard external memory architectures do not typically isolate the control and perceptual pathways. Instead, perceptual inputs are passed directly to a central controller, which is responsible for writing to and reading from a single, monolithic memory. Though it is possible for a role-filler structure to emerge in these systems given a sufficiently large amount of training data~\cite{chen2021learning}, they take much longer to learn relational tasks (requiring approximately an order of magnitude more training data), and do not generalize as well as the ESBN~\cite{webb2020emergent}. Thus, although external memory plays an important role in the ESBN, the presence of external memory alone is insufficient to enforce a relational bottleneck. Rather, it is the \textit{isolation} of the perceptual and abstract processing components from one another that does so. Furthermore, as we illustrate in the following sections, it is possible to achieve this isolation without the use of external memory.

\subsection*{Relation matrices}

An alternative approach to implementing the relational bottleneck is illustrated by the Compositional Relation Network (CoRelNet) (Figure~\ref{corelnet_arch})~\cite{kerg2022neural}. In that approach, a set of perceptual observations are first processed by an encoder, yielding a sequence of object embeddings. A relation matrix is then computed over all pairs of objects (within a given problem instance, e.g., all pairs of objects found within the ABA pattern in Figure~\ref{rel_bottleneck_illustrated}), in which each entry consists of the inner product between a pair of object embeddings (thus capturing the similarity between each pair of objects). Finally, this relation matrix is passed to a downstream decoder network (the architecture of this network can vary, e.g., using a multilayer perceptron or transformer). The relation matrix thus forms a relational bottleneck: all perceptual information flows into this matrix, converting it into a form that only preserves relational information (about the similarity between objects), and only this relational information is then passed on to the decoder. As with the ESBN, this relational bottleneck enables CoRelNet to rapidly learn and systematically generalize relational patterns. 

CoRelNet can be viewed as a feedforward, parallelized implementation of the sequential process (of encoding and similarity-based retrieval from external memory) carried out by the ESBN. This results in performance benefits, as CoRelNet does not suffer from the vanishing gradient problem that is a challenge for recurrent neural networks used to implement sequential processing~\cite{hochreiter1998vanishing}. That is, the gradients associated with all relations in a given scene are processed in parallel via a single step of backpropagation, rather than being diluted over several iterations, as is the case for the sequential processing carried out by the ESBN. This approach also makes the key relational inductive bias underlying the ESBN more explicit. The ESBN's memory retrieval procedure, in which the current observation is compared to the entries in memory, can be interpreted as computing a single row of the relation matrix. In both architectures, downstream processing is constrained so as to depend only on this relation matrix (which forms the relational bottleneck), though the details of this dependency differ.

Here too, a useful contrast can be made with related architectures that do not incorporate a relational bottleneck. In particular, architectures such as the Relation Net~\cite{santoro2017simple} (see~\cite{battaglia2018relational} for related approaches) explicitly perform a comparison between each pair of inputs, leading to improved performance in relational tasks. However, whereas CoRelNet represents pairwise relations using inner products, the Relation Net utilizes generic neural network components (e.g., multilayer perceptrons) that are learned in a task-dependent manner. While this is in principle more flexible, it does not constrain the network to learn representations that \textit{only} capture relational information. As a consequence, this architecture is susceptible to learning shortcuts consistent with the training data (i.e., overfitting to perceptual details), compromising its ability to efficiently learn and reliably generalize relations to out-of-distribution inputs~\cite{kim2018not,webb2020emergent,ichien2021visual}. This is in contrast to the inner product operation employed by the ESBN and CoRelNet, which is inherently relational, and therefore guarantees that downstream processing is based only on relations.

\subsection*{Relational attention}

The recently proposed Abstractor architecture (Figure~\ref{abstractor_arch})~\cite{altabaa2023abstractors} illustrates how the relational bottleneck can be implemented within the broader framework of attention-based architectures (including the Transformer~\cite{vaswani2017attention}). The Abstractor is built on a novel \textbf{attention} operation termed \textit{relational cross-attention}. In this operation, a set of object embeddings (which may be produced by an encoder given perceptual observations) is converted to form keys and queries, using separate linear projections. A relation matrix is then computed, in which each entry corresponds to the inner product between a query and key. The relation matrix is used to attend over a set of learned values. The attention operation itself is identical to standard self-attention, in which each value embedding is replaced with a weighted average of the set of all value embeddings, where the weights are determined by the match between queries and keys (i.e., the relation matrix). However, the value embeddings used for relational cross-attention are formed from a separate set of representations that are learned through backpropagation (based on the downstream task for which the network is trained) which \textit{reference} objects but are independent of their attributes (in the simplest scheme, these values reference objects via their position, though more sophisticated referencing schemes are also possible ~\cite{altabaa2023abstractors}). This is in contrast to standard self-attention, in which the values are computed via a linear projection of the perceptual inputs (just as the queries and keys are computed). As with the CoRelNet architecture, the relation matrix thus forms a relational bottleneck, in the sense that it converts all perceptual information to relational information, and downstream processing (in this case, the relational cross-attention operation) then depends only on this relational information.

Relational cross-attention can be contrasted with the standard forms of attention employed in Transformers: self-attention and cross-attention. In self-attention, the same set of object embeddings are used to generate keys, queries, and values. In cross-attention, object embeddings are used to generate keys and values, and queries are generated by a separate decoder network. In both cases, the values over which attention is performed are based directly on the object embeddings, and the information contained in these embeddings is therefore passed on for downstream processing (thus contravening the relational bottleneck). By contrast, in \textit{relational} cross-attention, keys and queries are generated from object embeddings, but a separate set of learned vectors are used as values (note that more complex architectures can be created by combining multiple forms of attention, e.g., object embeddings can first be processed by standard self-attention before applying relational cross-attention). As in the ESBN, these values can be viewed as learned `symbols', in the sense that they are isolated from the perceptual content of the objects with which they are associated.

This implementation of the relational bottleneck yields the same benefits observed in others: the Abstractor learns relational patterns faster than the Transformer, and displays better out-of-distribution generalization of those patterns. The Abstractor also has a few advantages relative to existing implementations of the ESBN and CoRelNet. Because the relation matrix is computed using separate key and query projections, the Abstractor is capable of representing asymmetric relations (e.g., can capture the difference in meanings between `A is greater than B' and `B is greater than A'; Figure~\ref{abstractor_results}). In addition, multi-head relational cross-attention enables the Abstractor to model multi-dimensional relations. As proposed, ESBN and CoRelNet are limited to relations along a single feature dimension only. Finally, similar to Transformers, the Abstractor is a \textit{generative} architecture, whereas the ESBN and CoRelNet are purely discriminative (although an alternative implementation of the ESBN has been proposed that can perform generative tasks~\cite{sinha2020memory}). This enables the Abstractor to perform a broader range of tasks, including the sequence-to-sequence tasks that are common in natural language processing.

\subsection*{Modeling more complex relations}

The neural network modeling work discussed in the previous sections was focused primarily, though not exclusively, on relational patterns involving same/different relations. Although similarity is fundamental to human reasoning, and has indeed been a central focus of theories of relational and analogical reasoning~\cite{falkenhainer1989structure,lu2022probabilistic,webb2022zero}, human cognition is also characterized by more complex relation types, including asymmetric relations~\cite{lu2019emergence} and higher-order relations~\cite{gentner1983structure} (relations between relations). It is worth emphasizing that the relational bottleneck can also account for many of these more complex relation types. First, although the inner product between two vectors is inherently symmetric ($\mathbf{a}\cdot\mathbf{b}$ is identical to $\mathbf{b}\cdot\mathbf{a}$), the relational bottleneck can also account for asymmetric relations by computing the inner product between separate key and query embeddings ($\mathbf{q_{a}}\cdot\mathbf{k_{b}}$ is not the same as $\mathbf{q_{b}}\cdot\mathbf{k_{a}}$). This use of separate key and query embeddings enables the Abstractor to model asymmetric relations such as greater-than/less-than, as well as other asymmetric relations found in mathematical reasoning problems~\cite{altabaa2023abstractors} (and a similar architectural modification is also possible for the ESBN and CoRelNet). Second, although the architectures discussed here focused on pairwise relations, higher-order relations can be straightforwardly accommodated through the recursive application of the relational bottleneck (i.e., by treating the outputs of one relational bottleneck as the inputs to another relational bottleneck, thus computing relations between other relations). Along these lines, recent work proposed relational convolutional networks~\cite{altabaa2023relational} and demonstrated that a hierarchical relational architecture can learn representations of higher-order relations, outperforming both non-hierarchical relational architectures (e.g., CoRelNet) and deep non-relational architectures (e.g., transformers). These results illustrate how the relational bottleneck can account for more complex relation types, but it remains an important avenue for future work to investigate whether and how such architectures can account for the full space of relations that characterize human cognition.

As the examples we have considered illustrate, the relational bottleneck can be implemented in a diverse range of architectures, each with their own strengths and weaknesses. In each case, the inclusion of a relational bottleneck enables rapid learning of relations without the need for pre-specified relational primitives. In the remainder of the review, we discuss the implications of this approach for models of cognition, and consider how the relational bottleneck may relate to the architecture of the human brain.

\section*{The relational bottleneck in the mind and brain}

\subsection*{Modeling the development of counting: a case study in learning abstractions}

A core requirement for cognitive models of abstract concept acquisition is to account for the timecourse of acquisition during human development. A useful case study can be found in the early childhood process of learning to count~\cite{wynn1992children,carey2001cognitive,sarnecka2008counting}. Children typically learn to recite the count sequence (i.e. `one, two, three,...' etc.) relatively early, but their ability to use this knowledge to count objects then proceeds in distinct stages (as measured by the `give N' task~\cite{sarnecka2008counting}, in which the child is asked to give the experimenter N objects). Each stage is characterized by the ability to reliably count sets up to a certain size (i.e., first acquiring the ability to reliably count only single objects, then to count two objects, and so on). Around the time that children learn to count sets of five, an inductive transition occurs, in which children rapidly learn to count sets of increasing size. It has been proposed that this transition corresponds to the acquisition of the `cardinality principle' -- the understanding that the last word used when counting corresponds to the number of items in a set~\cite{sarnecka2008counting} -- though the exact nature and scope of this inductive transition has been the subject of debate~\cite{davidson2012does,carey2019ontogenetic}. Previous work found that data accumulation played an essential role in supporting this inductive transition~\cite{piantadosi2012bootstrapping}, but that work employed a symbolic modeling approach, leaving open the question of how such symbol-like processes might be implemented in a neural system.

To address this, a recent study investigated the development of counting in deep neural network architectures~\cite{dulberg2021modelling}. These included the ESBN, the Transformer, and long short-term memory (LSTM)~\cite{hochreiter1997long} 
(a type of recurrent neural network). Each architecture displayed a distinct developmental timecourse (Figure~\ref{counting_results}). The Transformer displayed a roughly linear timecourse, taking approximately the same amount of time to master each number. The LSTM displayed an exponentially increasing timecourse, taking more time to learn each new number. Only the ESBN displayed a human-like inductive transition, gradually learning to count each number from one to four, and then rapidly acquiring the ability to count higher after learning to count to five. This was due to the ability of the ESBN to learn a procedure over the representations in its control pathway that was abstracted away from the specific numbers in the count sequence (represented in the model's perceptual pathway), allowing it to rapidly and systematically generalize between numbers. Specifically, the ESBN learned a procedure in which it stopped counting once the count sequence matched the desired number of objects (once the target value of N had been reached in the give-N task). The ESBN's use of symbol-like representations (in its control pathway, representing `item at which to stop’) allowed this procedure to be abstracted away from the particular target value (represented in the perceptual pathway, and bound to the symbol-like representation in external memory), facilitating rapid generalization to higher values. This case study illustrates how the relational bottleneck can facilitate a human-like developmental trajectory for learning abstract concepts. 

\subsection*{Capacity limits and the curse of compositionality}

The relational bottleneck principle may also help to explain the limited capacity of some cognitive processes (e.g., working memory)~\cite{miller1956magical}. Recent work has proposed a normative explanation of capacity-limited processes, according to which these capacity limits arise from the use of compositional representations, implemented in an architecture that employs a relational bottleneck~\cite{frankland2021no}. In that architecture, two separate representational pools (each representing distinct feature spaces, e.g., color and location) interact via a dynamic variable-binding mechanism (in that case, implemented using rapid Hebbian learning~\cite{hopfield1982neural}) in a manner that is conceptually similar to the ESBN. This mechanism enables the model to flexibly construct compositional representations (e.g., representing a visual scene by binding together spatial locations and visual features). However, this flexibility comes at the cost of relying on compositional representations that, by definition, are shared across many different, potentially competing processes. For instance, the representation of a blue object in the upper left corner (formed by binding together representations of `blue' and `upper left') will overlap with the representation of a blue object in the lower right corner (formed by binding together representations of `blue' and `lower right'), leading to interference between these two representations (note that this interference is also further exacerbated by the use of maximally low-dimensional codes to represent features). This can be viewed as an instance of the more general tradeoff between processing capacity and the use of shared representations~\cite{musslick2021rationalizing}, and also relates to classic considerations regarding the `binding problem' in psychology~\cite{treisman1980feature}, though it has not been previously appreciated that this tradeoff can explain the severely capacity-limited nature of processes such as working memory (and others, including subitizing~\cite{mandler1982subitizing}, and absolute judgment~\cite{pollack1952information}). This work illustrates how one of the most notable strengths of human cognition -- compositionality -- may explain one of its most notable weaknesses -- capacity-limited processing -- and how both can be implemented in neural networks via the relational bottleneck.



\noindent\rule{\textwidth}{1pt}
\subsubsection*{Box 3: Brain mechanisms supporting the relational bottleneck}

How might the relational bottleneck principle be implemented in the human brain? A central element of this framework is the presence of segregated systems for representing abstract vs. perceptual information (i.e., abstract values vs. perceptual keys/queries in the ESBN or Abstractor). A large body of findings from cognitive neuroscience suggests the presence of distinct neocortical systems for representing abstract structure (e.g., of space or events) vs. concrete entities (e.g., people or places), located in the parietal and temporal cortices respectively~\cite{mishkin1983object,goodale1992separate,frankland2020concepts,summerfield2020structure,o2022structure}. This factorization has also been explored in a number of recent computational models~\cite{russin2019compositional,o2021deep,bakhtiari2021functional}.

However, this segregation raises the question of how representations in these distinct neocortical systems are flexibly bound together. Though many proposals have been made for how the brain might solve this variable-binding problem (see Box 1), one intriguing possibility involves use of the episodic memory (EM) system~\cite{tulving2002episodic}. A common view holds that EM is supported by rapid synaptic plasticity in the hippocampus, which complements slower statistical learning in the neocortex~\cite{mcclelland1995there,sun2023organizing}. According to this view, episodes are encoded in the hippocampus by the rapid \textit{binding} of features that co-occur within an episode, while the features themselves are represented in neocortical systems. This same mechanism could in principle support an architecture similar to the ESBN, by enabling rapid binding of abstract and perceptual neocortical representations. This is in fact very similar to models of cognitive map learning, in which conjunctive representations are rapidly formed in the hippocampus~\cite{whittington2020tolman}. These hippocampal conjunctive codes bind structural and sensory information, which are thought to be encoded in the medial and lateral entorhinal cortices, respectively, and are commonly understood as extensions of the parietal and temporal neocortical systems referenced above. More generally, the involvement of EM in relational reasoning would be consistent with a growing body of recent findings suggesting the potential involvement of EM in tasks traditionally associated with working memory~\cite{hoskin2019refresh,beukers2021activity,beukers2023when}.

That said, the extent to which variable-binding relies on the hippocampus remains an open question. Some lesion evidence suggests that hippocampal damage does not lead to impairments of abstract reasoning~\cite{dzieciol2017hippocampal}. Other alternatives are that variable-binding may be supported by other structures capable of rapid synaptic plasticity (e.g., the cerebellum, which has been increasingly implicated in higher cognitive functions \cite{ravizza2006cerebellar,d2020evidence,mcdougle2022continuous}), or by other structures (such as the prefrontal cortex) that use other mechanisms for binding (such as selective attention~\cite{miller2001integrative} or working memory gating~\cite{kriete2013indirection}). The latter possibilities are consistent with findings that prefrontal damage often leads to severe deficits in abstract reasoning tasks~\cite{waltz1999system,cipolotti2023graph}, and prefrontal activity is frequently implicated in neuroimaging studies of abstract reasoning~\cite{christoff2001rostrolateral,knowlton2012neurocomputational}. However, this may also reflect the role of prefrontal cortex in \textit{representing} abstract structure (along with the parietal system described above), rather than the \textit{binding} of that structural information to concrete content. Of course, it is also possible that variable-binding is supported by a collection of distinct mechanisms, rather than a single mechanism alone. These are all important questions for future work that we hope will be usefully guided by the formalisms and computational models reviewed here.

\noindent\rule{\textwidth}{1pt}


\section*{Concluding remarks and future directions}

The human mind has a remarkable ability to acquire abstract relational concepts from relatively limited and concrete experience. Here, we have proposed the relational bottleneck as a functional principle that may explain how the human brain accomplishes such data-efficient abstraction, and highlighted recently proposed computational models that implement this principle. We have also considered how the principle relates to a range of cognitive phenomena, and how it might be implemented by the mechanisms of the human brain.

It should be noted that the framework reviewed here is not necessarily at odds with the existence of certain forms of domain-specific innate knowledge. In particular, a range of evidence from developmental psychology has suggested that humans possess certain `core knowledge' systems, such as an innate capacity to represent objects~\cite{spelke1992origins,spelke2007core,baillargeon2012core}. These findings have motivated the development of neuro-symbolic models endowed with these innate capacities~\cite{smith2019modeling}, although it is also possible that these findings may ultimately be accounted for by the inclusion of additional inductive biases into connectionist systems, such as mechanisms for object-centric visual processing~\cite{burgess2019monet,locatello2020object,piloto2022intuitive,mondal2023learning} (which have also been combined with the relational bottleneck~\cite{webb2023systematic}). Critically, however, it is important to emphasize that the relational bottleneck is, in principle, orthogonal to questions about these domain-specific capacities, and is focused instead on explaining the induction of abstract, domain-general concepts and relations.

There are a number of important avenues for further developing the relational bottleneck framework. One major question concerns how the framework relates to cognitive models of analogical reasoning, which have traditionally afforded a central role to the process of analogical \textit{mapping}, driven by patterns of similarity over entities and relations~\cite{gentner1983structure,falkenhainer1989structure,lu2022probabilistic,webb2022zero}. An intriguing possibility is that the relational bottleneck encourages neural networks to learn to implement such algorithms, by re-representing their inputs in terms of patterns of similarity (represented as inner products), though future work should aim to more precisely establish this link. Future work should also consider how the proposed framework relates to other theoretical perspectives on out-of-distribution generalization in neural networks, including group theoretic approaches~\cite{bronstein2021geometric,segert2024symmetry}, as well as other cognitive processes relevant to abstraction, including attentional processes~\cite{vaishnav2022gamr} and semantic cognition~\cite{giallanza2023integrated}. Additionally, much work has suggested that human reasoning is not purely relational, but instead depends on a mixture of concrete and abstract influences~\cite{wason1968reasoning,johnson1972reasoning,bassok1998adding,goldberg2003constructions}. This suggests the potential value of a more graded formulation that controls the amount of non-relational information allowed to pass through the bottleneck. Finally, the human capacity for abstraction surely depends not only on architectural biases such as those that we have discussed here, but also on the rich educational and cultural fabric that allows us to build on the abstractions developed by others~\cite{mcclelland2022capturing}. In future work, it will be important to explore the interaction between education, culture and relational inductive biases.


\section*{Outstanding Questions}

\begin{itemize}
    \item Can a more graded version of the relational bottleneck capture `content effects’ -- in which abstract reasoning processes are influenced by the specific content under consideration, and therefore are not purely abstract or relational -- while preserving a capacity for relational abstraction?
    \item Can the relational bottleneck principle be usefully applied to symbolic (or neuro-symbolic) models, in a manner similar to its application to neural network models? For instance, could program induction models benefit from a constraint that forces perceptual inputs to be recoded in terms of relations?
    \item What is the relationship between the relational bottleneck and traditional cognitive models of analogical reasoning? Does the relational bottleneck provide a useful inductive bias toward learning to implement processes such as analogical mapping?
    \item How can other cognitive processes (attention, memory, etc.) be integrated with the relational bottleneck?
    \item How is the relational bottleneck implemented in the brain? To what extent does this rely on mechanisms responsible for episodic memory, attentional mechanisms, and/or other mechanisms that remain to be identified? What role do the hippocampus, prefrontal cortex, and/or other structures play in these computations?
    \item How do architectural biases toward relational processing interact with the influence of training curricula and other cultural sources of abstraction (e.g., formal education)?
\end{itemize}

\section*{Glossary}

\subsubsection*{Attention}

In the context of deep learning, an operation that allows information to be flexibly shared between a set of embeddings~\cite{bahdanau2014neural} (popularized by the Transformer architecture~\cite{vaswani2017attention}), also closely related to content-addressable memory (see External memory). This operation shares some, but not all, properties of formal treatments of attention in cognitive psychology (see~\cite{cohen1990control} for an example of how the psychological notion of attention may be implemented in neural networks; see~\cite{lindsay2020attention} for discussion of the relationship between these different senses of ‘attention’).

\subsubsection*{Backpropagation}

A technique used to train multi-layer neural networks, in which the connection strengths between processing units in intermediate layers are updated based on an error signal in downstream layers, allowing intermediate representations to be automatically learned in a manner that is most useful for downstream tasks (without having to specify those intermediate representations).

\subsubsection*{Connectionism}

A modeling framework in cognitive science that emphasizes the emergence of complex cognitive phenomena from the interaction of simple, neuron-like elements organized into networks, in which connections are formed through learning.

\subsubsection*{Embedding}

A real-valued vector that represents a particular input (e.g., a word or image), often instantiated as the state (set of activation values) of a particular layer in a neural network.

\subsubsection*{Episodic memory}

A form of memory in which arbitrary, but durable, associations can be rapidly formed. Often thought to be implemented by hippocampal mechanisms for rapid synaptic plasticity and similarity-based retrieval.

\subsubsection*{External memory}

In the context of deep learning, an approach that combines neural networks with separate external stores of information, typically with learnable mechanisms for writing to and reading from these stores, and in which retrieval is usually similarity-based (i.e., `content-addressable'). Often used to implement a form of episodic memory~\cite{medin1978context}.

\subsubsection*{Inductive bias}

An assumption made by a machine learning model about the distribution of the data. In deep learning models, this often takes the form of architectural features that bias learning toward certain (typically desirable) outcomes. Genetically pre-configured aspects of brain structure can be viewed as a form of inductive bias.

\subsubsection*{Inner product}

An operation in which two vectors are converted to a scalar, often interpreted as representing the similarity of those two vectors.

\subsubsection*{Out-of-distribution generalization}

In machine learning, generalization to a distribution that differs from the distribution observed during training.

\subsubsection*{Program induction}

A modeling approach in which concepts are represented as symbolic programs and learned via a search process that seeks to maximize the likelihood of observed data (typically subject to some constraints, e.g., parsimony).

\section*{Declaration of interests}

The authors declare no competing interests.

\section*{Acknowledgements}

AA is supported by funds provided by the National Science Foundation and by DoD OUSD (R\&E) under Cooperative Agreement PHY-2229929 (The NSF AI Institute for Artificial and Natural Intelligence). JDC is supported by Vannevar Bush Faculty Fellowship N00014-22-1-2002 from the Office of the Under Secretary of Defense for Research \& Engineering, supported by ONR.


\printbibliography

\end{document}